\documentclass[11pt,twoside]{article}
\usepackage{jmlr2e}

\usepackage[utf8]{inputenc} % allow utf-8 input
\usepackage[T1]{fontenc}    % use 8-bit T1 fonts
\usepackage{siunitx}
\usepackage{multirow}
\usepackage{url}            % simple URL typesetting
\usepackage{booktabs}       % professional-quality tables
\usepackage{amsfonts}       % blackboard math symbols
\usepackage{nicefrac}       % compact symbols for 1/2, etc.
\usepackage{microtype}      % microtypography
\usepackage{floatrow}

\usepackage{amsmath}
\usepackage{amssymb}
\usepackage{txfonts}

\renewcommand{\nu}{m} % this “renames” \nu to a different variable name

\usepackage{lastpage}
\jmlrheading{23}{2022}{1-\pageref{LastPage}}{1/21; Revised 12/21}{3/22}{21-0055}{Jan Niklas Böhm, Philipp Berens, and Dmitry Kobak}
\ShortHeadings{Attraction and Repulsion in Neighbor Embeddings}{Böhm, Berens, and Kobak}
\firstpageno{1}
\editor{Samuel Kaski}

\title{Attraction-Repulsion Spectrum in Neighbor Embeddings}

\author{%
  Jan Niklas B\"ohm \email{jan-niklas.boehm@uni-tuebingen.de}
  \AND
  Philipp Berens   \email philipp.berens@uni-tuebingen.de
  \AND
  Dmitry Kobak \email dmitry.kobak@uni-tuebingen.de\\
  \addr Institute for Ophthalmic Research, University of T\"ubingen\\ Otfried-Müller-Str. 25; 72076 Tübingen, Germany
}

\begin{document}

\maketitle

\begin{abstract}%
Neighbor embeddings are a family of methods for visualizing complex high-dimensional data sets using $k$NN graphs. To find the low-dimensional embedding, these algorithms combine an attractive force between neighboring pairs of points with a repulsive force between all points. One of the most popular examples of such algorithms is t-SNE. Here we empirically show that changing the balance between the attractive and the repulsive forces in t-SNE using the exaggeration parameter yields a spectrum of embeddings, which is characterized by a simple trade-off: stronger attraction can better represent continuous manifold structures, while stronger repulsion can better represent discrete cluster structures and yields higher $k$NN recall. We find that UMAP embeddings correspond to t-SNE with increased attraction; mathematical analysis shows that this is because the negative sampling optimization strategy employed by UMAP strongly lowers the effective repulsion. Likewise, ForceAtlas2, commonly used for visualizing developmental single-cell transcriptomic data, yields embeddings corresponding to t-SNE with the attraction increased even more. At the extreme of this spectrum lie Laplacian eigenmaps. Our results demonstrate that many prominent neighbor embedding algorithms can be placed onto the attraction-repulsion spectrum, and highlight the inherent trade-offs between them. 
\end{abstract}
\section{Introduction}

\begin{figure}[t]
  \centering
  \includegraphics[width=\linewidth]{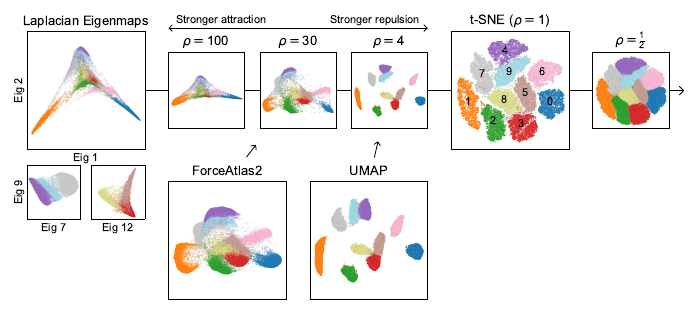}
  \caption{\textbf{Attraction-repulsion spectrum for the MNIST data.} Different embeddings of the MNIST data set of hand-written digits ($n=70\,000$); colors denote digits as shown in the t-SNE panel. Multiplying all attractive forces by an exaggeration factor $\rho$ yields a spectrum of embeddings. Values below 1 yield inflated clusters. Values above 1 yield more compact clusters. Higher values make multiple clusters merge, with $\rho\to\infty$ approximately corresponding to Laplacian eigenmaps \citep{linderman2019clustering}. Insets show two subsets of digits separated in higher eigenvectors. UMAP is similar to $\rho\approx 4$. ForceAtlas2 is similar to $\rho\approx 30$.}
  \label{fig:spectrum}
\end{figure}

T-distributed stochastic neighbor embedding (t-SNE) \citep{maaten2008visualizing} is arguably among the most popular methods for low-dimensional visualization of complex high-dimensional data sets. It defines pairwise similarities called \textit{affinities} between points in the high-dimensional space and aims to arrange the points in a low-dimensional space to match these affinities \citep{hinton2003stochastic}. Affinities decay exponentially with high-dimensional distance, making them infinitesimal for most pairs of points and making the $n\times n$ affinity matrix effectively sparse. Efficient implementations of t-SNE  \citep{maaten2014accelerating, linderman2019fast} explicitly truncate the affinities and use the $k$-nearest-neighbor ($k$NN) graph of the data with $k\ll n$ as the input.  

We use the term \textit{neighbor embedding} (NE) to refer to all dimensionality reduction methods that operate on the $k$NN graph of the data and aim to preserve neighborhood relationships \citep{yang2013scalable, yang2014optimization}. A prominent recent example of this class of algorithms is UMAP \citep{mcinnes2018umap}, which has become popular in applied fields such as single-cell transcriptomics \citep{becht2019dimensionality}. It is based on stochastic optimization and typically produces more compact clusters than t-SNE.

Another example of neighbor embeddings are force-directed graph layouts \citep{noack2007energy, noack2009modularity}, originally developed for graph drawing. One specific algorithm called ForceAtlas2 \citep{jacomy2014forceatlas2} has recently gained popularity in the single-cell transcriptomic community to visualize data sets capturing cells at different stages of development 
\citep{weinreb2018spring, weinreb2020lineage, wagner2018single, tusi2018population, kanton2019organoid, sharma2020emergence}.

In general, NE algorithms optimize the layout using attractive forces between all pairs of points connected by a $k$NN graph edge, thus placing them closer in the low-dimensional embedding.  In addition, every point feels a repulsive force to every other point, which prevents trivial solutions, such as positioning all points on top of each other.  While earlier algorithms took inspiration from physical systems \citep{fruchterman1991}, similar concepts arise naturally from the loss functions grounded in information theory (see below).

Here we  provide a unifying account of NE algorithms. We study the spectrum of t-SNE embeddings that are obtained when increasing/decreasing the attractive forces between $k$NN graph neighbors, thereby changing the balance between attraction and repulsion. This leads to a trade-off between faithful representations of continuous and discrete structures. Remarkably, we discover that ForceAtlas2 and UMAP can both be accurately positioned on this spectrum (Figure~\ref{fig:spectrum}). For UMAP, we use mathematical analysis and Barnes--Hut re-implementation to show that increased attraction is due to the negative sampling optimization strategy, and to derive the effective repulsion strength. 

All our code is available at \url{https://github.com/berenslab/ne-spectrum}.

\section{Related Work}

Various trade-offs in SNE and t-SNE generalizations \citep{yang2009heavy, carreira2010elastic, kobak2019heavy,venna2010information,amid2015optimizing, amid2019trimap, narayan2015alpha, im2018stochastic} as well as in graph layout algorithms \citep{noack2007energy, gansner2012maxent} have been studied previously, but our work is the first to study the \textit{exaggeration}-induced trade-off in t-SNE. Prior work used `early exaggeration' only as an optimization trick \citep{maaten2008visualizing} that allows to separate well-defined clusters \citep{linderman2019clustering, arora2018analysis}.

\citet{carreira2010elastic} introduced the \textit{elastic embedding} algorithm that has an explicit parameter $\lambda$ controlling the attraction-repulsion balance. However, that paper suggests slowly increasing $\lambda$ during optimization, as an optimization trick similar to the early exaggeration, and does not discuss trade-offs between high and low values of $\lambda$. The same holds for the graph layout model studied by \citet{gansner2012maxent}, who similarly anneal the repulsion strength during optimization.

Our results on UMAP go against the common wisdom regarding what makes UMAP perform as it does \citep{mcinnes2018umap, becht2019dimensionality}. No previous work suggested that negative sampling may have a drastic effect on the resulting embedding. In a follow-up paper, \citet{damrich2021umap} analyzed the effective loss function of UMAP in more detail.

\section{Neighbor Embeddings}

The standard expositions of t-SNE, UMAP, and ForceAtlas2 (FA2) create the impression that these algorithms have little to do with each other. They use different affinities, different loss functions, different optimization strategies, and different large-sample approximations. They are introduced using different motivations. Importantly, the loss function in t-SNE includes a normalizing term which makes its optimization difficult, whereas the loss functions of UMAP and FA2 do not have such a term.

Despite all these differences, we claim that these algorithms are intimately related (Figure \ref{fig:spectrum}). In this section, we cast t-SNE, UMAP, FA2, and Laplacian eigenmaps (LE) in a common mathematical framework, using consistent notation and highlighting the similarities between them. The empirical results will be presented in the following sections. We denote the original high-dimensional points as $\mathbf x_i$ and their low-dimensional positions as~$\mathbf y_i$.

\subsection{T-SNE}

T-SNE measures similarities between $\mathbf x_i$ by \textit{affinities} $v_{ij}$ and \textit{normalized affinities} $p_{ij}$ 
\begin{equation*}
    p_{ij} = \frac{v_{ij}}{n},\;\;\;
    v_{ij} = \frac{p_{i|j} + p_{j|i}}{2},\;\;\;
    p_{j|i} = \frac{v_{j|i}}{\sum_{k\ne i} v_{k|i}},\;\;\;
    v_{j|i}=\exp\left(-\frac{\|\mathbf x_i - \mathbf x_j \|^2}{2\sigma_i^2}\right).
\end{equation*}
For fixed $i$, $p_{j|i}$ is a probability distribution over all points $j\ne i$ (all $p_{i|i}$ are set to zero), and the variance of the Gaussian kernel $\sigma_i^2$ is chosen to yield a pre-specified value of the \textit{perplexity} of this probability distribution, $\mathcal P=2^\mathcal H$, where $\mathcal H=-\sum_{j\ne i} p_{j|i} \log_2 p_{j|i}$ is the entropy. The symmetrized affinities $v_{ij}$ are then normalized  by $n$  for $p_{ij}$ to form a probability distribution over the set of all pairs of points $(i,j)$. Modern implementations \citep{maaten2014accelerating, linderman2019fast} construct a $k$NN graph with $k=3\mathcal P$ neighbors and only consider affinities between connected nodes as non-zero. The default perplexity value in most implementations is $\mathcal P=30$.

\begin{figure}
    \centering
    \floatbox[{\capbeside\thisfloatsetup{capbesideposition={right,top},capbesidewidth=0.55   \textwidth}}]{figure}[\FBwidth]
    {\caption{\label{fig:variant-tsne}{\textbf{The role of affinities in t-SNE.}} MNIST data set. \textbf{(a)}~Default t-SNE, Gaussian affinities, perplexity 30. \textbf{(b)}~t-SNE with binary $k$NN affinities: all nonzero $p_{ij}$ are the same, and $p_{ij}>0$ iff point $i$ is among 15 nearest neighbors of point $j$, or vice versa.}}
    {\includegraphics{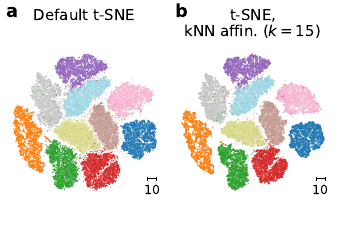}}
\end{figure}

While t-SNE traditionally uses Gaussian affinities, the affinity matrix can be simplified without having a large impact on the resulting layout. In particular, one can use the $k$NN ($k=15$) adjacency matrix $\mathbf A = [a_{ij}]$ to construct symmetric binary affinities $v_{ij} = a_{ij} \vee a_{ji}$, and then obtain $p_{ij}$ by normalizing the entire matrix to sum to 1. The resulting `$k$NN affinities' typically yield t-SNE embeddings that are almost identical to the default ones (Figure~\ref{fig:variant-tsne}).

Similarities in the low-dimensional space are defined as
\begin{equation*}
    q_{ij} = \frac{w_{ij}}{Z},\;\;\;
    w_{ij} = \frac{1}{1 + d_{ij}^2},\;\;\;
    d_{ij} = \|\mathbf y_i - \mathbf y_j\|,\;\;\;
    Z=\sum_{k\ne l} w_{kl},
\end{equation*}
with all $q_{ii}$ set to 0. The points $\mathbf y_i$ are then rearranged in order to minimize the Kullback--Leibler (KL) divergence $\mathcal D_\mathrm{KL}\big(\{p_{ij}\}\,\|\,\{q_{ij}\}\big) = \sum_{i,j} p_{ij}\log\big(p_{ij}/q_{ij}\big)$ between $p_{ij}$ and $q_{ij}$
\begin{align}
    \mathcal L_\textrm{t-SNE} \sim  -\sum_{i,j} p_{ij}\log \frac{w_{ij}}{Z} =-\sum_{i,j} p_{ij}\log w_{ij} + \log \sum_{i,j}w_{ij},
\end{align}
where we drop constant terms and take into account that $\sum p_{ij}=1$. The first term  contributes attractive forces to the gradient while the second term yields repulsive forces. Indeed, using $\partial w_{ij}/\partial \mathbf y_i= -2w^2_{ij}(\mathbf y_i-\mathbf y_j)$, the gradient, up to a constant factor, can be written as
\begin{align}
    \frac{\partial \mathcal L_\textrm{t-SNE}}{\partial \mathbf y_i}
    &\sim \sum_j v_{ij}w_{ij}(\mathbf y_i-\mathbf y_j) - \frac{n}{Z}\sum_j w_{ij}^2(\mathbf y_i-\mathbf y_j).
    \label{eq:tsneGradient}
\end{align}

\subsection{Exaggeration in t-SNE}
\label{sec:tSNEexaggeration}

A standard optimization trick for t-SNE called \textit{early exaggeration} \citep{maaten2008visualizing,maaten2014accelerating} is to multiply the first sum in the gradient by a factor $\rho>1$ during the initial iterations of gradient descent. This increases the attractive forces and allows similar points to gather into clusters more effectively. Modern implementations use $\rho=12$ for the initial 250 iterations \citep{maaten2014accelerating} by default. The gradient of t-SNE with exaggeration can be written as 
\begin{equation}
    \frac{\partial \mathcal L_\textrm{t-SNE}(\rho)}{\partial \mathbf y_i} \sim \sum_j v_{ij}w_{ij}(\mathbf y_i-\mathbf y_j) - \frac{n}{\rho Z}\sum_j w_{ij}^2(\mathbf y_i-\mathbf y_j)
    \label{eq:tsneGradientExagg}
\end{equation}
and the corresponding loss function can be written in a functional form akin to the KL divergence
\begin{equation}
    \mathcal L_\textrm{t-SNE}(\rho) 
    = \sum_{i,j} p_{ij}\log\frac{p_{ij}}{w_{ij}/Z^\frac{1}{\rho}}.
    \label{eq:tsne-exag}
\end{equation}
However, for $\rho\ne 1$ the values $w_{ij}/Z^\frac{1}{\rho}$ in the denominator do not sum to 1 so Eq.~\eqref{eq:tsne-exag} is not a KL divergence between probability distributions.

\subsection{UMAP}

\begin{figure}
    \centering
    \includegraphics{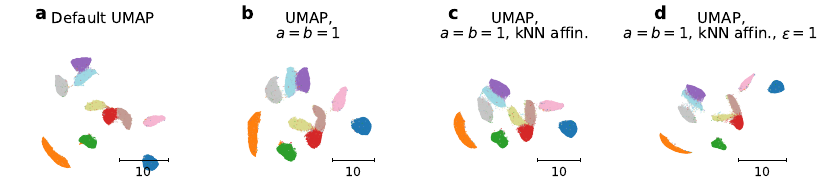}
    \caption{\label{fig:variant-umap}\textbf{UMAP with various simplifications.} MNIST data set. \textbf{(a)} Default UMAP with $a\approx1.6$ and $b\approx 0.9$ and LE initialization. \textbf{(b)} UMAP with $a=b=1$ and PCA initialization, the default choice for our experiments. \textbf{(c)}~The same as in (b), but using binary $k$NN affinities ($v_{ij} = 1$ iff point $i$ is among $15$ nearest neigbors of point $j$, or vice versa). \textbf{(d)} The same as in (c), but with $\epsilon=1$.}
\end{figure}

Using the same notation as above, UMAP aims to optimize the cross-entropy loss between $v_{ij}$ and $w_{ij}$, without normalizing them into probabilities:
\begin{equation}
    \mathcal L_\mathrm{UMAP} = \sum_{i,j} \left[v_{ij}\log\frac{v_{ij}}{w_{ij}} + (1-v_{ij})\log\frac{1-v_{ij}}{1-w_{ij}}\right],
    \label{eq:umapLoss}
\end{equation}
where the $1-v_{ij}$ term is approximated by $1$ as most $v_{ij}$ are 0, as done in the original implementation of UMAP \citep[see][Section~2.3.2]{sainburg2021parametric}. Note that UMAP differs from t-SNE in how exactly it defines $v_{ij}$ (it uses adaptive Laplacian kernel with $k=15$ by default), but its result does not change much when using the same binary affinities $v_{ij}$ we introduced above for t-SNE (Figure~\ref{fig:variant-umap}c). Therefore, we believe that the difference in affinities is not what drives the difference in layout between t-SNE and UMAP in practice; see below for the experimental evidence.

Dropping terms that do not depend on $\mathbf{y}_i$, we obtain
\begin{equation}
    \mathcal L_\mathrm{UMAP} \sim - \sum_{i,j}v_{ij}\log w_{ij} - \sum_{i,j} \log(1-w_{ij}),
\end{equation}
which is the same loss function as the one introduced earlier in LargeVis \citep{tang2016visualizing}. 
The first term, corresponding to attractive forces, is the same as in t-SNE, but the second, repulsive, term is different. Taking $w_{ij}=1/(1+d_{ij}^2)$ as in t-SNE,
the UMAP gradient is given by
\begin{equation}
     \frac{\partial \mathcal L_\mathrm{UMAP}}{\partial \mathbf y_i} \sim \sum_j v_{ij}w_{ij}(\mathbf y_i-\mathbf y_j) - \sum_j \frac{1}{d_{ij}^2 +\epsilon}w_{ij}(\mathbf y_i-\mathbf y_j),
     \label{eq:umapGradient}
\end{equation}
where $\epsilon=0.001$ is added to the denominator to prevent numerical problems for $d_{ij}\approx 0$. 
Note that UMAP uses  $w_{ij}=1/(1+ad^{2b}_{ij})$ as an output kernel with $a\approx1.6$ and $b\approx 0.9$ by default. However, setting $a=b=1$ does not strongly affect the result (Figure~\ref{fig:variant-umap}b). Moreover, when we modified the UMAP implementation to set $\epsilon=1$, the resulting embeddings also stayed qualitatively similar (Figure~\ref{fig:variant-umap}d). 
So here again, we believe that these details are not what drives the difference in layout between t-SNE and UMAP in practice; see below for the experimental evidence.

If $\epsilon=1$, the gradient becomes identical to the t-SNE gradient, up to the $n/Z$ factor in front of the repulsive forces. Moreover, UMAP implementation allows to use an arbitrary $\gamma$ factor in front of the repulsive forces, which makes it easier to compare the loss functions
\begin{equation}
     \frac{\partial \mathcal L_\mathrm{UMAP}(\gamma)}{\partial \mathbf y_i} \sim \sum_j v_{ij}w_{ij}(\mathbf y_i-\mathbf y_j) - \gamma \sum_j \frac{1}{d_{ij}^2 +\epsilon}w_{ij}(\mathbf y_i-\mathbf y_j).
     \label{eq:umapGradientGamma}
\end{equation}
Note that LargeVis used $\gamma=7$ by default but UMAP sets $\gamma=1$, as follows from its cross-entropy loss function.

Whereas it is possible to approximate the full repulsive term with the same techniques as used in t-SNE \citep{maaten2014accelerating, linderman2019fast}, UMAP takes a different approach and follows LargeVis in using \textit{negative sampling} \citep{mikolov2013distributed} of repulsive forces: on each gradient descent iteration, only a small number $\nu$ of randomly picked repulsive forces are applied to each point for each of the ${\sim}k$ attractive forces that it feels. Other repulsive terms are ignored. 
The default value is $\nu=5$. The effect of negative sampling on the resulting embedding has not been studied before.

\subsection{ForceAtlas2}

\begin{figure}
    \centering
    \floatbox[{\capbeside\thisfloatsetup{capbesideposition={right,top},capbesidewidth=0.5\textwidth}}]{figure}[\FBwidth]
    {\caption{\label{fig:variant-fa2} \textbf{The effect of edge repulsion in Force\-Atlas2.} MNIST data set. \textbf{(a)} FA2 with repulsion by degree. \textbf{(b)}~FA2 without repulsion by degree.  Note the difference in scale.}}
    {\includegraphics{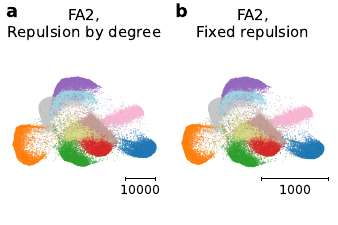}}
\end{figure}

Force-directed graph layouts are usually introduced directly via attractive and repulsive forces, even though it is easy to write down a suitable loss function \citep{noack2007energy}. ForceAtlas2 (FA2) has attractive forces proportional to $d_{ij}$ and repulsive forces proportional to $1/d_{ij}$ \citep{jacomy2014forceatlas2}:
\begin{equation}
    \frac{\partial \mathcal L_\mathrm{FA2}}{\partial \mathbf y_i} = \sum_j v_{ij}(\mathbf y_i-\mathbf y_j) - \sum_j \frac{(h_i+1)(h_j+1)}{d_{ij}^2}(\mathbf y_i-\mathbf y_j),
    \label{eq:FA2gradient}
\end{equation}
where $h_i$ denotes the degree of node $i$ in the input graph. This is known as \textit{edge repulsion} in the graph layout literature \citep{noack2007energy, noack2009modularity} and is important for embedding graphs that have nodes of very different degrees. However, for symmetrized $k$NN graphs, %\footnote{If the points are distributed uniformly on the Riemannian manifold (same assumption as UMAP) then the neighbors of a given point will have the given point as a neighbor, too.  As such, the symmetrization process will not change the number of $k$ nearest neighbors significantly.} 
assuming that they do not have too many `hubs' \citep{radovanovic2010hubs}, $h_i\approx k$, so $(h_i+1)(h_j+1)$ term contributes a roughly constant ${\sim}k^2$ factor to the repulsive forces, and can be compensated by decreasing all distances by a factor of $k$. Indeed, for the MNIST data set, removing the edge repulsion factor led to a ${\sim}15$ times decrease in scale but otherwise almost the same embedding (Figure~\ref{fig:variant-fa2}). 

\subsection{Laplacian Eigenmaps}
\label{sec:laplacian-eigenmaps}

Laplacian eigenmaps \citep{belkin2002laplacian,coifman2006diffusion} is a method for dimensionality reduction that leverages spectral graph theory.  Its loss function can be written with a quadratic constraint
\begin{equation}
    \mathcal L_\textrm{LE} = \sum_{ij} v_{ij} \|\mathbf y_i-\mathbf y_j\|^2\;\;\mathrm{s.\,t.}\;\;\mathbf Y^\top \mathbf D \mathbf Y = \mathbf I,
\end{equation}
where $\mathbf D$ is a diagonal matrix with $D_{ii} = \sum_j V_{ij}$ for affinity matrix $\mathbf V=[v_{ij}]$, $\mathbf I$ is the identity matrix, and $\mathbf Y$ is the embedding matrix having $\mathbf y_i$ as rows. This loss function can be minimized by solving a generalized eigenvalue problem (Appendix A). The quadratic constraint in some sense serves the role of repulsive forces, preventing collapse of the embedding to a single point.

\citet{carreira2010elastic} and \citet{linderman2019clustering} noticed that the attractive term in the t-SNE loss function on its own reduces to the unconstrained loss function of Laplacian eigenmaps. Indeed, if $\rho\to\infty$, the relative repulsion strength becomes infinitesimal and the embedding shrinks to a point with all $w_{ij}\to 1$. This means that the gradient from Equation~\ref{eq:tsneGradient} reduces to $\sum_j v_{ij} (\mathbf y_i-\mathbf y_j)$, which coincides with the  gradient of Laplacian eigenmaps (apart from the quadratic constraint).
%This implies that, asymptotically, gradient descent becomes equivalent to Markov chain iterations with the transition matrix closely related to the graph Laplacian $\mathbf L = \mathbf D - \mathbf V$ of the affinity matrix $\mathbf V=[v_{ij}]$ (here $\mathbf D$ is diagonal matrix with row sums of $\mathbf V$; see Appendix). 
A more detailed analysis in Appendix A shows that when $\rho\to\infty$, the entire embedding shrinks to a single point, but the leading eigenvectors of the graph Laplacian $\mathbf L = \mathbf D - \mathbf V$ shrink the slowest. This makes t-SNE with large values of $\rho$ produce embeddings very similar to LE, which computes the leading eigenvectors of the normalized Laplacian (Appendix A). 

This theoretical finding immediately suggests that it might be interesting to study t-SNE with exaggeration $\rho>1$ not only as an optimization trick, but in itself, as an intermediate method between LE and standard t-SNE.

\subsection{Implementation}
\label{sec:implementation}

All experiments were performed in Python. We ran all packages with default parameters, unless specified otherwise. We used openTSNE 0.6.0 \citep{policar2019opentsne}, a Python reimplementation of FIt-SNE \citep{linderman2019fast}. When using $\rho<12$, we used the default early exaggeration with $\rho_\mathrm{early}=12$, and exaggeration $\rho$ for all subsequent iterations. For $\rho\ge 12$ no early exaggeration was used and exaggeration $\rho$ was applied throughout. The learning rate was set to $\eta=n/\max(\rho,\rho_\mathrm{early})$ as suggested by \citet{belkina2019automated}. Note that we used default Gaussian affinities for all experiments.

We used UMAP 0.5.1 with Cauchy similarity kernel (by setting $a=b=1$). We used default UMAP affinities for all experiments.  The Barnes--Hut implementation of UMAP was developed in Cython \citep{behnel2010cython}, on top of the \texttt{openTSNE} package.  We extended the package to leave out the calculation of the normalization constant~$Z$, take into account the $\epsilon$ and $\gamma$ parameters from Equation~\ref{eq:umapGradientGamma}, and load the default UMAP affinities as computed by UMAP itself. For these experiments we also set $a=b=1$.

For FA2 we used the \texttt{fa2} package \citep{fa2}, which employs a Barnes--Hut approximation to speed up computation of the repulsive forces.  We developed a patch that makes it possible to disable the repulsion by degree and applied it on top of the current version 0.3.5.  The input to FA2 was the unweighted symmetrized approximate $k$NN graph $\mathbf A \vee \mathbf A^\top$, where $\mathbf A$ is the $k$NN adjacency matrix constructed with \texttt{Annoy} \citep{annoy} with $k=15$.  By default, all algorithms were optimized for 750 iterations.

Unless stated otherwise, we used principal component analysis (PCA) initialization to remove any differences between algorithms due to initialization strategies \citep{kobak2019umap} and to make all embeddings of the same data set visually aligned to each other \citep{kobak2018art}. For t-SNE, the initialization was always scaled to have a standard deviation of  0.0001, as suggested by \citet{kobak2018art} and is default in \texttt{openTSNE} \citep{policar2019opentsne}.  For UMAP, the initialization  was scaled to have the range of $[-10, 10]$, as is default in the original implementation. For ForceAtlas2, we scaled the initialization to have a standard deviation of $10\thinspace000$ to approximately match the scale of final ForceAtlas2 embeddings (we experimented with different values and found this setting to work well and avoid convergence problems). Note that Figure~\ref{fig:toydata} is an exception and uses random initialization. By default, UMAP uses LE (and not PCA) initialization, whereas the \texttt{fa2} implementation uses random initialization. On the data sets analyzed here, we did not observe any qualitative difference as a result of our non-default (PCA) initialization (cf. Figure~\ref{fig:variant-umap}a,b).

LE was computed using the \texttt{scikit-learn} \citep{pedregosa2011scikit} implementation, using the \texttt{SpectralEmbedding} class and the \texttt{PyAMG} solver~4.0 \citep{olson2018pyamg}. The input graph was the same as the input to FA2. No initialization was needed for LE. We flipped the signs of LE eigenvectors to orient them similarly to other embeddings, whenever necessary.

\section{The Attraction-Repulsion Spectrum}

\begin{figure}[t]
  \centering
  \includegraphics[]{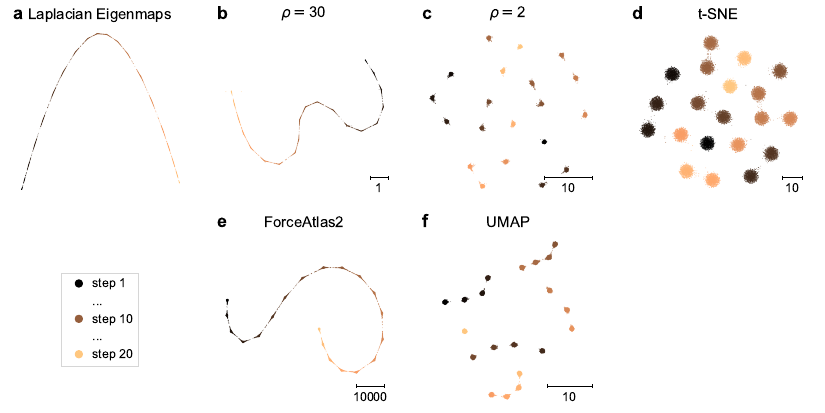}
  \caption{\textbf{Simulated data emulating a developmental trajectory.}  The points were sampled from 20 isotropic 50-dimensional Gaussians, equally spaced along one axis such that only few inter-cluster edges exist in the $k$NN graph. Panels (b--f) used a shared random initialization. Panels (b--d) did not use early exaggeration.}
  \label{fig:toydata}
\end{figure}

We first investigated the relationships between the NE algorithms using the MNIST data set of hand-written digits (sample size $n=70\,000$; dimensionality $28\times28=784$, reduced to 50 with PCA; Figure~\ref{fig:spectrum}). T-SNE produced an embedding where all ten digits were clearly separated into clusters with little white space between them, making it difficult to assess relationships between digits. Increasing attraction to $\rho=4$ shrank the clusters and strongly increased the amount of white space; it also identified two groups of graphically similar digits: ``4/7/9'' and ``3/5/8''. Further increasing the attraction to $\rho=30$ made all clusters connect together: cluster ``6'' connected to ``5'' and to ``0''. Even higher exaggeration made the embedding similar to Laplacian eigenmaps, in agreement with the theoretical prediction discussed above \citep{linderman2019clustering}. Here similar digit groups like ``4/7/9'' were entirely overlapping, and could only be separated using higher eigenvectors (Figure~\ref{fig:spectrum}, insets).  On the other side of the spectrum, exaggeration values $0<\rho<1$ resulted in inflated coalescing clusters.

The MNIST example suggests that high attraction emphasizes connections between clusters at the cost of within-cluster structure, whereas high repulsion emphasizes the cluster structure at the expense of between-cluster connections. We interpreted this finding as a \textit{continuity-discreteness trade-off}.

We developed a simple toy example to illustrate this trade-off in more detail (Figure~\ref{fig:toydata}). For this, we generated data as draws from 20 standard isotropic Gaussians in 50 dimensions, each shifted by $6$ standard deviation units from the previous one along one axis (1000 points per Gaussian, so $n=20\,000$ overall). For this analysis we used random initialization and turned the early exaggeration off, to isolate the effect of each loss function on the `unwrapping' of the random initial configuration.

We found that t-SNE with strong exaggeration ($\rho=30$) recovered the underlying one-dimensional manifold structure of the data almost as well as LE (Figure~\ref{fig:toydata}a,b). In both cases, the individual clusters were almost invisible. In contrast, an embedding with weaker attraction and stronger repulsion (t-SNE with exaggeration $\rho=2$) showed individual clusters but was unable to fully recover the 1-dimensional structure and only found some chunks of it (Figure~\ref{fig:toydata}c). Finally, standard t-SNE clearly showed 20 individual clusters but with the continuous structure entirely lost (Figure~\ref{fig:toydata}d).

\begin{figure}[t]
  \centering
  \includegraphics[width=\linewidth]{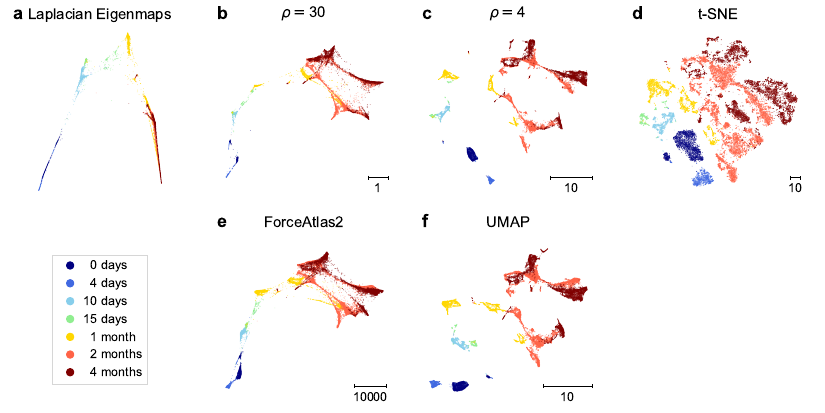}
  \caption{\textbf{Neighbor embeddings of the single-cell RNA-seq developmental data.} Cells were sampled from human brain organoids (cell line \texttt{409b2}) at seven time points between 0 days and 4 months into the development \citep{kanton2019organoid}. Sample size $n=20\,272$. Data were reduced with PCA to 50 dimensions. See Appendix B for transcriptomic data preprocessing steps.}
  \label{fig:treutleinHuman}
\end{figure}

Further, we analyzed a developmental single-cell transcriptomic data set, where cells were collected from human brain organoids at seven time points between 0 days and 4 months into the development \citep{kanton2019organoid}. In this kind of data, one expects to find rich cluster structure as well as a strong time-dependent trajectory. As in the other data sets, we found that stronger attraction ($\rho=30$) better represented the developmental trajectory, whereas stronger repulsion (standard t-SNE) better represented the cluster structure (Figure~\ref{fig:treutleinHuman}). Using much higher $k$ for the $k$NN graph construction made the developmental trajectory in high-attraction methods even clearer (Figure~\ref{fig:treutleinHumanHighK}), in agreement with the FA2-based analysis performed in the original publication. We observed the same pattern in a separate data set obtained from chimpanzee brain organoids (Figures~\ref{fig:treutleinChimp}, \ref{fig:treutleinChimpHighK}).

\begin{figure}
    \centering
    \floatbox[{\capbeside\thisfloatsetup{capbesideposition={right,center},capbesidewidth=0.4\textwidth}}]{figure}[\FBwidth]
    {\includegraphics{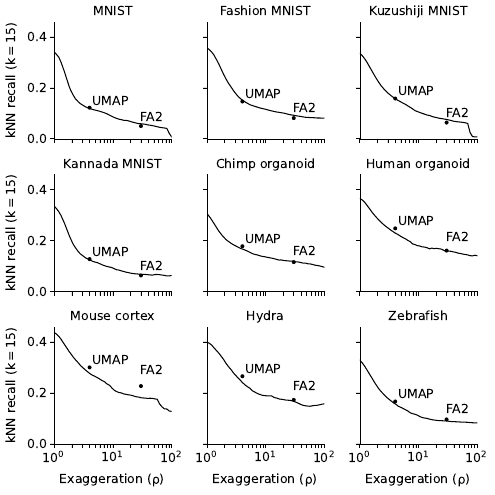}}
    {\caption{\label{fig:knn-recall}\textbf{Nearest neighbors recall as a function of $\rho$.}  The fraction of $k=15$ nearest neighbors in high dimensions that remain among the nearest neighbors in the embedding (average over $10\thinspace000$ randomly selected points; see text). The values for UMAP and FA2 are shown at $\rho=4$ and $\rho=30$.}}
\end{figure}

While high exaggeration helps to preserve continuous  structures, this comes with a price of distorting local neighborhoods. To quantify this effect, we computed the fraction of $k=15$ nearest neighbors in high dimensions that remain among the nearest neighbors in the embedding (`$k$NN recall'). To compute it for a given data point, we found 15 points with the largest affinities in the symmetrized affinity matrix, and determined what fraction of them is among the 15 exact nearest neighbors in the embedding. This was averaged over $10\thinspace000$ randomly selected points. We found that as $\rho$ increased, the local neighborhood became more and more distorted (Figure~\ref{fig:knn-recall}). For the MNIST data set, the $k$NN recall of default t-SNE ($\rho=1$) was 0.34; with $\rho=4$ it went down to 0.12; with $\rho=30$ it further dropped to 0.06.

We observed the same fast and monotonic decrease in $k$NN recall in both brain organoid data sets, as well as in six further data sets (Figure~\ref{fig:knn-recall}): Fashion MNIST \citep{xiao2017fashion}, Kannada MNIST \citep{prabhu2019kannada}, Kuzushiji MNIST \citep{clanuwat2018kuzmnist}, single-cell data from hydra \citep{siebert19hydra}, from zebrafish embryo \citep{wagner2018zfish}, and from mouse cortex \citep{tasic2018shared}.

\section{UMAP and ForceAtlas2 Can Be Placed on the Attraction-Repulsion Spectrum}

Using the MNIST data set, we observed that FA2 produced an embedding very similar to t-SNE with $\rho\approx30$, while UMAP produced an embedding very similar to t-SNE with $\rho\approx4$ (Figure~\ref{fig:spectrum}). The same was true for the simulated data set (Figure~\ref{fig:toydata}) and for the brain organoid data set (Figure~\ref{fig:treutleinHuman}), as well as for the seven further data sets that we analyzed in addition (Figures~\ref{fig:treutleinChimp}, \ref{fig:faminst}, \ref{fig:kannada}, \ref{fig:kuzmnist}, \ref{fig:hydra}, \ref{fig:zfish}, \ref{fig:tasic}). 

\begin{figure}
    \centering
    \includegraphics{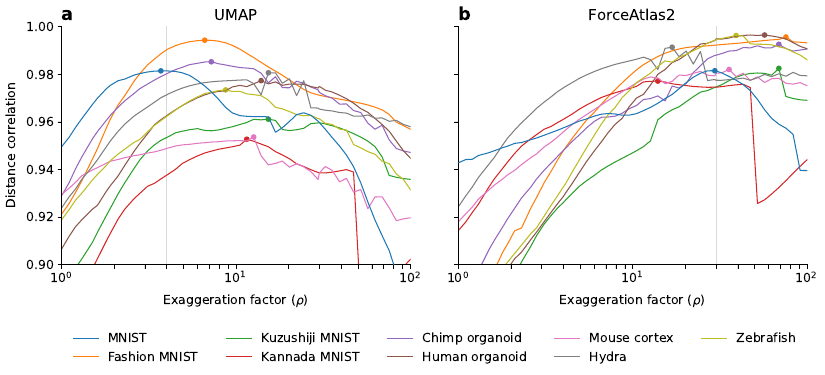}
    \caption{\textbf{Distance correlations between UMAP/FA2 and t-SNE.} Exaggeration values $\rho\in[1, 100]$ were evenly distributed on a log-scale, with $\rho=4$ and $\rho=30$ added explicitly; 52 points in total.  Distance correlation \citep{szekely2007measuring} was computed using \texttt{dcor} package \citep{dcor} on a random subset ($n=5\,000$) of the data. Dots mark the maximum of each curve.  \textbf{(a)} Distance correlation between UMAP and t-SNE. \textbf{(b)}~Distance correlation between FA2 and t-SNE.}
    \label{fig:results-dist-corr}
\end{figure}

To quantify this observation, we computed distance correlations \citep{szekely2007measuring} between UMAP \& FA2 embeddings and t-SNE embeddings with various values of $\rho\in[1,100]$ (Figure~\ref{fig:results-dist-corr}). We found that for most data sets the highest correlation between UMAP and t-SNE layouts was achieved at $4 \le \rho < 15$ (Figure~\ref{fig:results-dist-corr}a). For FA2, the highest correlation was typically achieved at $20 < \rho < 80$ (Figure~\ref{fig:results-dist-corr}b). In both cases, the maximum correlations were above 0.94, indicating very similar layouts. Whereas the exact value of $\rho$ yielding the maximum correlation varied between data sets, the correlation values at $\rho=4$ for UMAP and at $\rho=30$ for FA2 were always high and close to the maximum correlations. We also observed that $\rho=4$  yielded a good match to UMAP independent of the sample size (Figure~\ref{fig:layoutCorrelationsEtc}).

Note that  for all three algorithms we used all default parameters (apart from always using the same PCA initialization and fixing $a=b=1$ in UMAP), confirming that the differences between t-SNE and UMAP in affinities and in the value of $\epsilon$ in the loss function do not play a large role, at least for our data sets. We used the shared PCA initialization to roughly align the embeddings and make distance correlations more meaningful, but all conclusions stayed the same if we used default UMAP/FA2 initializations (see Section~\ref{sec:implementation}). 

A caveat here is that distance correlation metric can be strongly affected by the exact placement of the clusters, and does not always capture the intuitive notion of `similarity'. For example, both correlation curves for the Kannada~MNIST data set (Figure \ref{fig:results-dist-corr}, red lines) peak at almost the same value of $\rho$, but visual inspection of the embeddings (Figure~\ref{fig:kannada}) suggests that $\rho=4$ is qualitatively closer to UMAP, while $\rho=30$ is qualitatively closer to FA2, in agreement with all other data sets.

The $k$NN recall of UMAP and FA2 was also similar to the $k$NN recall of t-SNE with exaggeration set to $\rho=4$ and $\rho=30$ respectively, across all analyzed data sets (Figure~\ref{fig:knn-recall}). This suggests that not only the general layout, as measured by the distance correlation, but also the local structure of the embedding was similar between UMAP/FA2 and t-SNE with appropriate exaggeration.

\section{Increased Attraction in UMAP Due to Negative Sampling}

\begin{figure}
    \centering
    \includegraphics{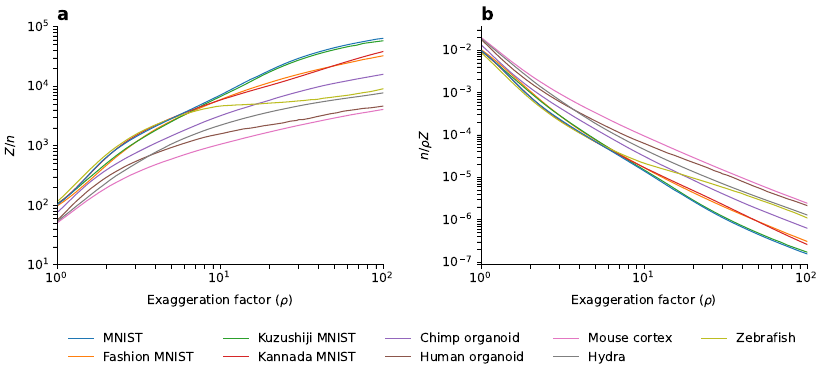}
    \caption{\textbf{(a)} The normalization term $Z/n$ computed for all data sets considered in the manuscript, as a function of~$\rho$.  \textbf{(b)}~The term $n/(\rho Z)$ computed for all data sets considered in the manuscript.}
    \label{fig:all-n-rho-Z}
\end{figure}

As shown above, the gradient of UMAP (Eq.~\ref{eq:umapGradient}) is very similar to the gradient of t-SNE (Eq.~\ref{eq:tsneGradient}) but does not contain the `normalizing' $n/Z$ term in front of the repulsive forces. What are the typical values of this coefficient? The normalization term $Z$ in t-SNE evolves during optimization: it starts at $Z\approx n^2$ due to all $d_{ij}\approx 0$ at initialization and decreases as the embedding expands. For a perfect embedding with all $p_{ij}=q_{ij}$ and $v_{ij}=w_{ij}$, $Z$ would be equal to $n$; in reality $Z$ usually still exceeds $n$. We found that for all nine data sets analyzed here, the value of $Z$ in the end of optimization with $\rho=1$ was in the range $[50n, 120n]$ (Figure~\ref{fig:all-n-rho-Z}a).  For MNIST, the final $Z$ value was ${\sim}100 n$, corresponding to the final $n/Z\approx 0.01$ (Figure~\ref{fig:all-n-rho-Z}b). Increasing the exaggeration shrinks the embedding and increases the final $Z$; it also changes the repulsive factor to $n/(\rho Z)$ (Eq.~\ref{eq:tsneGradientExagg}). Across all data sets, the final $Z$ value with $\rho=4$ was in the $[400n,2300n]$ range (Figure~\ref{fig:all-n-rho-Z}a). For MNIST, it was ${\sim}2100 n$, corresponding to the final $n/(\rho Z)\approx 0.0001$. This means that UMAP matched t-SNE results with the repulsive factor $0.0001$ better than it matched t-SNE results with the repulsive factor $0.01$, even though UMAP itself uses repulsive factor $\gamma=1$ (Eq.~\ref{eq:umapGradient}). How is this possible?

\begin{figure}[t]
  \centering
  \includegraphics[width=\linewidth]{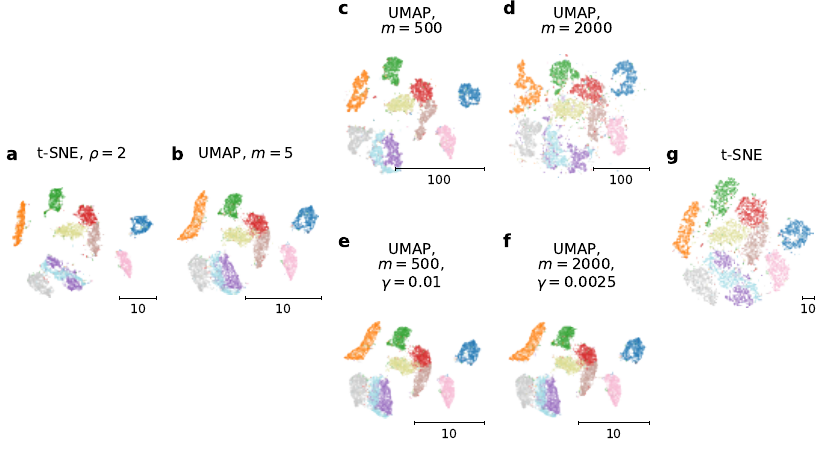}
  \caption{\textbf{The effect of negative sampling rate on UMAP embeddings.} MNIST subsample with $n=6000$. We used a subsample of MNIST because the runtime scales as $\mathcal O(\nu n)$, making it impractical to use $\nu\approx n$ for large $n$. UMAP (panels b--f) was run for 3000 epochs to ensure convergence, as large $m$ introduce noise in the optimization.  \textbf{(a)} T-SNE embedding with $\rho=2$. \textbf{(b--d)} UMAP embeddings with $\nu\in\{5,500,2000\}$.  Panels (c) and (d) were initialized with the standard UMAP embedding ($\nu=5$, 750 epochs) to simulate early exaggeration. 
  \textbf{(e--f)} UMAP embeddings with $\nu\in\{500,2000\}$, while keeping the product $\gamma\cdot\nu$ constant.
  \textbf{(g)} Standard t-SNE of the same data.}
  \label{fig:umapRepulsion}
\end{figure}

We hypothesized that this mismatch arises because the UMAP implementation is based on negative sampling and does not in fact optimize its stated loss (Eq.~\ref{eq:umapLoss}). Instead, the negative sampling decreases the repulsion strength, creating an effective $\gamma_\mathrm{eff}(\nu) \ll 1$. We verified that increasing the value of $\nu$ increased the repulsion strength in UMAP (Figure~\ref{fig:umapRepulsion}): embeddings grew in size and the amount of between-cluster white space decreased. But when we decreased the $\gamma$ factor together with increasing $\nu$ so that their product $\gamma\cdot\nu$ stayed constant, the embedding did not change at all (Figure~\ref{fig:umapRepulsion}e,f), confirming that the negative sampling rate $\nu$ directly controls the repulsion strength. 

The repulsion strength in UMAP can also be explicitly controlled by the $\gamma$ parameter. Decreasing the $\gamma$ value had the same effect as increasing the $\rho$ value in t-SNE, and moved the UMAP result towards the LE part of the attraction-repulsion spectrum (Figure~\ref{fig:umapAttraction}). We found it not possible to increase the repulsion strength by setting $\gamma \gg 1$, likely due to convergence problems.

We can approximately compute the effective repulsion coefficient $\gamma_\mathrm{eff}(\nu)$ arising in UMAP through the negative sampling as follows. 
The number of repulsive forces per one attractive force is ${\sim}n/k$ in the full gradient but $m$ with negative sampling. This suggests that $\gamma_\mathrm{eff}\approx km/n$ and should decrease with the sample size as $\mathcal O(1/n)$.

\begin{figure}[t]
  \centering
  \includegraphics[width=\linewidth]{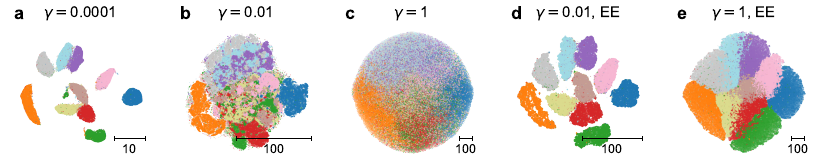}
  \caption{\textbf{Barnes--Hut UMAP without negative sampling.} \textbf{(a--c)} Embeddings with $\gamma\in\{0.0001, 0.01, 1\}$. \textbf{(d--e)} Embeddings with gamma values $\gamma\in\{0.01, 1\}$ initialized with the embedding with $\gamma=0.0001$ [panel (a)], in analogy to early exaggeration in t-SNE.}
  \label{fig:bhumap}
\end{figure}

To confirm our interpretation, we developed a Barnes--Hut UMAP implementation that optimizes the full UMAP loss without any negative sampling (see Section~\ref{sec:implementation}). On full MNIST, $\gamma=0.0001$ yielded an embedding that resembled the standard (negative-sampling-based) UMAP (Figure~\ref{fig:bhumap}a), while larger values of $\gamma$ yielded over-repulsed embeddings (Figure~\ref{fig:bhumap}b,c) and required early exaggeration to produce meaningful results (Figure~\ref{fig:bhumap}d,e), with $\gamma=0.01$ resembling t-SNE and $\gamma=1$ being over-repulsed compared to t-SNE. This suggests that directly optimizing the cross-entropy loss (Eq.~\ref{eq:umapLoss}) leads to an embedding where the repulsive forces strongly dominate visual appearance (Figure~\ref{fig:bhumap}c,e).

\begin{figure}
    \centering
    \floatbox[{\capbeside\thisfloatsetup{capbesideposition={right,top},capbesidewidth=0.5\textwidth}}]{figure}[\FBwidth]
    {\caption{\textbf{Estimating the effective repulsion in UMAP.} We ran Barnes--Hut UMAP for subsets of MNIST with sample sizes from $10\,000$ to $70\,000$ in the increments of $5000$, using a grid of $\gamma$ values. For each sample size, we found the $\gamma$ value matching the size of negative-sampling-based UMAP. The grey line shows the linear regression fit to the log-transformed data. Sample sizes below $10\,000$ led to noisy estimates and were excluded.}
    \label{fig:gamma-effective}}
    {\includegraphics{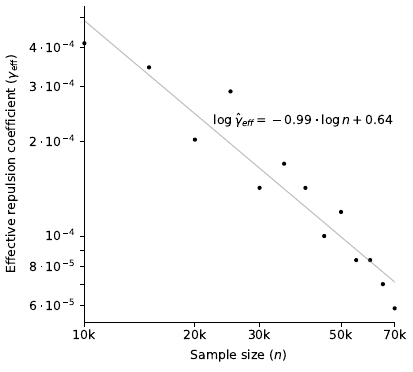}}
    
\end{figure}

Furthermore, we can use our Barnes--Hut implementation to directly estimate $\gamma_\mathrm{eff}$. To do that, we ran the Barnes--Hut implementation for a range of sample sizes and a range of $\gamma$ values (40 values of $\gamma$ evenly log-spaced between $0.01$ and $10^{-5}$). For each sample size, we found the $\gamma$ value giving the best match to the size ($\max \mathbf{Y} - \min \mathbf{Y}$) of the negative-sampling-based UMAP (Figure~\ref{fig:gamma-effective}). The resulting $\hat\gamma_\mathrm{eff}$ estimates decreased with the sample size as $1/n$, with regression slope being $-0.99$ on the log-log plot (Figure~\ref{fig:gamma-effective}), confirming our prediction. In a follow-up work, \citet{damrich2021umap} analyzed the UMAP sampling procedure in more detail, showing that $\gamma_\mathrm{eff}\approx \log k \cdot m/n$.

In popular expositions \citep{coenen2019understanding,oskolkov2019how}, the success of UMAP and its visually appealing embeddings have been attributed to its cross-entropy loss function and its topological foundations. However, our conclusion is that the more condensed clusters typically observed in UMAP compared to t-SNE are a serendipitous consequence of UMAP's negative sampling strategy. The mathematical framework and the cross-entropy loss function developed in the original paper \citep{mcinnes2018umap} would lead to very different and suboptimal embeddings (Figure~\ref{fig:bhumap}c,e), if not for the negative sampling. Note that we are not criticizing the actual UMAP embeddings here; our statement is that UMAP's stated and effective loss functions are qualitatively different.

\section{Increased Attraction in FA2 Due to Non-Decaying Attractive Forces}

The attractive forces in t-SNE scale as $d_{ij}/(1+d_{ij}^2)$. When all $d_{ij}$ are small, this becomes an approximately linear dependency on $d_{ij}$, which is the reason why t-SNE with high exaggeration $\rho \gg 1$ replicates Laplacian eigenmaps (see Section~\ref{sec:laplacian-eigenmaps} and Appendix~A). For large distances $d_{ij}$, attractive forces in t-SNE decay to zero, making default t-SNE very different from LE.  In contrast, in FA2, attractive forces always scale as $d_{ij}$. Thus, the larger the embedding distance between two points, the stronger the attractive force between them. This strong non-decaying attractive force moves FA2 towards Laplacian eigenmaps on the attraction-repulsion spectrum.

While the attractive forces in FA2 are the same as in Laplacian eigenmaps, FA2 has repulsive forces instead of the quadratic constraint of LE. This moves FA2 somewhat away from LE on the attraction-repulsion spectrum. %As such, FA2 only reaches an equilibrium once the scale of the embedding grows large enough to accommodate the repulsive force, while at the same time drawing adjacent pairs in high-dimensional space together with an ever-growing force.
These arguments provide a qualitative explanation for why FA2 behaves similar to t-SNE with strong exaggeration ($\rho\approx 30$, as we empirically showed above), but more quantitative analysis remains for future work. In addition, our arguments suggest that the exact scaling law of the repulsive forces ($1/d_{ij}^2$ or $1/d_{ij}$) may have little qualitative influence on the resulting embedding as long as the attractive forces remain linear in $d_{ij}$. We leave it for future work to investigate this.

Note that it is not possible to move FA2 embeddings along the attraction-repulsion spectrum by multiplying the attractive or repulsive forces by a constant factor (such as $\gamma$ in UMAP or $\rho$ in t-SNE). Multiplying attractive forces by any factor $a$ or repulsive forces by any factor $1/a$ only leads to rescaling of the embedding by $1/\sqrt{a}$. Indeed, if all forces are in equilibrium before such multiplication and rescaling, they will stay in equilibrium afterwards. This is a general property of force-directed layouts where both attractive and repulsive forces scale as powers of the embedding distance $d_{ij}$.

\section{Discussion}

We showed that changing the balance between attractive and repulsive forces in t-SNE directly affects the trade-off between preserving continuous/global or discrete/local structures. Increasingly strong repulsion `brings out' information from higher Laplacian eigenvectors into the two embedding dimensions (Figure~\ref{fig:spectrum}). It is remarkable that the repulsive forces, which are data-agnostic and do not depend on the input data \citep{carreira2010elastic}, have so much qualitative influence.

While we only considered the exaggeration factor $\rho$ here, other parameters of t-SNE can also qualitatively affect the resulting embedding. In particular, the tail-heaviness of the low-dimensional similarity kernel \citep{yang2009heavy} controls the emphasis put on the fine cluster structure of the data \citep{kobak2019heavy}. There is thus non-trivial interaction between the exaggeration $\rho$ and the tail-heaviness parameter  $\alpha$, which we illustrate using the MNIST data set in  Figure~\ref{fig:tsne-tail-heavy}, but leave more detailed exploration of this two-dimensional parameter space for future work.

Our results suggest that it is beneficial for high-repulsion embeddings to  begin optimization with lower repulsion strength, in order to better preserve global structure. This explains how UMAP benefits from its default initialization with Laplacian eigenmaps \citep{kobak2019umap} and how t-SNE benefits from early exaggeration \citep{linderman2019clustering} (Figure~\ref{fig:tsneParams} demonstrates the importance of early exaggeration in t-SNE). Similarly, \citet{carreira2010elastic} suggested to gradually increase repulsion strength during optimization of {elastic embedding}. A promising approach to t-SNE optimization would be to use Laplacian eigenmaps for initialization and replace the early exaggeration phase with gradual annealing of the exaggeration factor $\rho$ from `infinity' down to its final desired value. 

Our treatment provides a unified perspective on several well-known NE algorithms that have scalable implementations and that have been shown to successfully embed data sets such as MNIST without coarse-graining the $k$NN graph. Methods based on coarse-graining, such as PHATE \citep{moon2019visualizing} or latent variable NE method in \citet{saul2020tractable} may behave differently. We believe that our treatment may allow to position other NE algorithms on the same spectrum. For example, a recently suggested TriMap algorithm \citep{amid2019trimap}, which uses negative sampling similar to UMAP, appears to have stronger attractive forces than UMAP (cf. Figure 5 in the original paper), with some TriMap embeddings, e.g. of the Fashion MNIST data set, looking similar to the ForceAtlas2 embeddings shown in our work.
It remains for future work to investigate if and how some of the more recent NE algorithms based on negative sampling fit on the attraction-repulsion spectrum. This includes, for example, IHVD  \citep{minch2020gpu} and MDE  \citep{agrawal2021minimum}. The latter work developed a flexible NE framework that can combine various attractive and repulsive forces optimized using negative sampling, with the quadratic constraint of Laplacian eigenmaps, resulting in a rich family of embeddings.

We argue that negative sampling \citep{mikolov2013distributed}, used by LargeVis/UMAP, strongly lowers the effective repulsion, compared to the stated cross-entropy loss function. In a follow-up work, \citet{damrich2021umap} have developed a more formal analysis of negative sampling in UMAP and confirmed our findings.

Negative sampling exhibits some similarity to stochastic gradient descent (SGD), where the gradient is repeatedly computed on small random subsets of the data, known as mini-batches. However, we believe that this analogy is not helpful. SGD iterates over the entire training set, partitioned in mini-batches. Small mini-batches increase the variance of the gradient estimates but do not introduce any bias. Negative sampling, on the other hand, only samples a small subset of the repulsive forces for each attractive force, introducing a systematic bias into the gradient computation.

Negative sampling is closely related to the  \textit{noise-contrastive estimation} (NCE) framework \citep{gutmann12nce}. NCE was recently applied to t-SNE under the name of NCVis \citep{artemenkov2020ncvis}, and the general NCE theory asserts that it should be asymptotically equivalent to optimizing the full gradient \citep{gutmann12nce}. We consider it an interesting research direction to study the relationship between negative sampling and NCE and their effect on 2D embeddings as well as on higher-dimensional embeddings used in methods like \texttt{word2vec} \citep{mikolov2013distributed}.

The practical takeaway from our work is not that one of the considered algorithms is the `best'. All three algorithms discussed in this manuscript (t-SNE, UMAP, ForceAtlas2) are widely used in several academic fields, like single-cell biology \citep{becht2019dimensionality, kobak2018art} or population genomics \citep{diaz2019umap, karczewski2020mutational}. But the choice of the method is often done without a solid understanding of why the results may be different or what trade-offs are at play. Our work highlights that which algorithm is more appropriate may depend on the question one wants to answer.

If the goal of the analysis is to explore the fine clusters and the local neighborhood structure, one should use t-SNE with default exaggeration parameter $\rho=1$ (Figure~\ref{fig:knn-recall}).
If the interest rather lies in exploring the global structure of the data, such as continuous or temporal trajectories, then it may be helpful to increase the exaggeration in t-SNE or to use UMAP or ForceAtlas2. All things considered, t-SNE with exaggeration is able to cover the entire attraction-repulsion spectrum (Figure~\ref{fig:spectrum}), whereas UMAP with negative sampling is only able to produce visualizations with relatively high attraction. ForceAtlas2 is even more limited in that it cannot move along the spectrum and corresponds to high attraction. In terms of the running speed, modern implementations of t-SNE \citep{linderman2019fast, artemenkov2020ncvis} and UMAP are comparable.

We hope that the treatment developed here will allow researchers to make an informed choice between algorithms in practical applications.

\goodbreak
\acks{%
We thank Sebastian Damrich, George Linderman, Stefan Steinerberger, James Melville, and Ulrike von Luxburg for helpful discussions; Pavlin Poli\v{c}ar for discussions and \texttt{openTSNE} support; and He Zhisong for the help with loading the organoid transcriptomic data.

This research was funded by the Cyber Valley Research Fund (D.30.28739), the Deutsche Forschungsgemeinschaft through a Heisenberg Professorship (BE5601/4-1), the Excellence Cluster 2064 ``Machine Learning: New Perspectives for Science'' (390727645), the National Institute Of Mental Health of the National Institutes of Health (U19MH114830), and the German Ministry of Education and Research (01IS18039A, 01GQ1601). The content is solely the responsibility of the authors and does not necessarily represent the official views of the National Institutes of Health. The authors declare no financial conflict of interest and did not receive any donations/funding from industry with relationship to this project.

The authors thank the International Max Planck Research School for Intelligent Systems (IMPRS-IS)
for supporting Jan Niklas Böhm.
}

\section*{Appendix A. Relationship to Laplacian Eigenmaps}

\textit{Laplacian eigenmaps}\quad Let a $n\times n$ symmetric matrix $\mathbf V$ contain pairwise affinities between $n$ points (or edge weights between nodes in an undirected graph). Let the diagonal matrix $\mathbf D$ contain row (or, equivalently, column) sums of $\mathbf V$,  that is, $D_{ii} = \sum_j V_{ij}$. Then $\mathbf L=\mathbf D-\mathbf V$ is known as the (unnormalized) graph Laplacian. Laplacian eigenmaps \citep{belkin2002laplacian} can then be formulated as solving the generalized eigenvector problem
\begin{equation*}
    \mathbf L\mathbf a = \lambda\mathbf D \mathbf a
\end{equation*}
and taking the eigenvectors corresponding to the \textit{smallest} eigenvalues (after discarding the trivial eigenvector $[1,1,\ldots, 1]^\top$ with eigenvalue zero). By multiplying both sides of this equation by $\mathbf D^{-1}$, the problem can be reformulated as finding the eigenvectors of $\mathbf D^{-1}\mathbf V$ corresponding to the \textit{largest} eigenvectors:
\begin{equation*}
    \mathbf D^{-1} \mathbf V \mathbf a = (1-\lambda)\mathbf a.
\end{equation*}
The matrix $\mathbf D^{-1}\mathbf V$ is not symmetric and has rows normalized to $1$. It can be interpreted as a diffusion operator on the graph, making Laplacian eigenmaps equivalent to \textit{Diffusion maps} \citep{coifman2006diffusion}. Another equivalent way to rewrite it, is to define normalized Laplacian $\mathbf L_\mathrm{norm} = \mathbf D^{-1/2} \mathbf L \mathbf D^{-1/2}$ and solve an eigenvector problem $\mathbf L_\mathrm{norm} \mathbf b = \lambda \mathbf b$, where $\mathbf b = \mathbf D^{
1/2} \mathbf a$.

\medskip\noindent
\textit{t-SNE without repulsion}\quad In the limit of $\rho\to\infty$, the repulsive term in the t-SNE gradient can be dropped, all $w_{ij}\to 1$, and hence the gradient descent update rule becomes \citep{linderman2019clustering}
\begin{equation*}
    \mathbf y_i^{t+1} = \mathbf y_i^t - \eta \sum_j v_{ij}(\mathbf y_i^t - \mathbf y_j^t),
\end{equation*}
where $t$ indexes the iteration number  and $\eta$ is the learning rate (including all constant factors in the gradient). Denoting by $\mathbf Y$ the $n\times 2$ matrix of the embedding coordinates, this can be rewritten as
\begin{align*}
    \mathbf Y^{t+1}& = (\mathbf I - \eta \mathbf D + \eta \mathbf V) \mathbf Y^t\\
    &= \mathbf M \mathbf Y^t.
\end{align*}
$\mathbf M$ is the transition matrix of this Markov chain (note that it is symmetric and its rows and columns sum to 1; its values are all non-negative for small enough $\eta$). According to the general theory of Markov chains, the largest eigenvalue of $\mathbf M$ is 1, and the corresponding eigenvector is $[1,1,\ldots, 1]^\top$, meaning that the embedding shrinks to a single point (as expected without repulsion). The slowest shrinking eigenvectors correspond to the next eigenvalues. This means that when $\rho$ becomes very large, the embedding will resemble leading nontrivial eigenvectors of $\mathbf M$ (as $\rho$ grows, the embedding will become smaller and smaller, but here we ignore its overall size; eigenvectors can have arbitrary length so overall scale of the embedding is not important here). This becomes equivalent to a power iteration algorithm. The eigenvectors of $\mathbf M$ are the same as of $\mathbf L = \mathbf D - \mathbf V$, which is the unnormalized graph Laplacian of the symmetric affinity matrix.

Note that this is not precisely what LE computes: as explained above, LE finds eigenvectors of the \textit{normalized} graph Laplacian \citep{luxburg2008consistency}. However, for t-SNE affinities, $\mathbf D$ is  approximately proportional to the identity matrix, because $\mathbf V$ is obtained via symmetrization of directed affinities, and those have rows summing to 1 by construction. We can therefore expect that the leading eigenvectors of $\mathbf L$ and of $\mathbf L_\mathrm{norm}$ are very close. We verified that for MNIST data they were almost exactly the same. A more abstract perspective is provided by \citet{luxburg2008consistency}: the unnormalized spectral embedding corresponds to minimizing a graph cut with respect to the number of vertices, whereas LE takes vertex degrees into account. Since all vertices in a $k$NN or perplexity-based graph have an almost equal degree, the eigenvectors of the normalized and unnormalized Laplacian will be very close.

Note also that nothing prevents different columns of $\mathbf Y$ to converge to the same leading eigenvector: each column independently follows its Markov chain. Indeed, we observed that for large enough values of $\rho$ and large enough number of gradient descent iterations, the embedding collapsed to one dimension. This is the expected limiting behaviour when $\rho\to\infty$. However, for moderate values of $\rho$ (as  shown in this manuscript), this typically does not happen, and columns of $\mathbf Y$ resemble the two leading non-trivial eigenvectors of the Laplacian. The repulsive force prevents the embedding from collapsing to the leading Laplacian eigenvector. At the same time, a weak repulsive force will only be able to `bring out' the second LE eigenvector. The stronger the contribution of repulsive forces, the more LE eigenvectors it would be able to `bring out' (remember that the attractive force acts stronger on the higher eigenvectors).

\medskip\noindent
\textit{Loss function of LE and quadratic constraint}\quad The original Laplacian eigenmaps paper \citep{belkin2002laplacian} motivates the eigenvector problem by considering
\begin{equation*}
    \mathcal L_\mathrm{LE} = \sum_{i,j} v_{ij} \|\mathbf y_i - \mathbf y_j\|^2 = 2\operatorname{Tr}(\mathbf Y^\top \mathbf L \mathbf Y).
\end{equation*}
This expression can be trivially minimized by setting all $\mathbf y_i = \mathbf 0$, so the authors introduce a quadratic constraint $\mathbf Y^\top \mathbf D \mathbf Y = \mathbf I$, yielding the generalized eigenvector problem. We note that a different quadratic constraint $\mathbf Y^\top \mathbf Y = \mathbf I$ would yield a simple eigenvector problem for $\mathbf L$. In any case, the constraint plays the role of the repulsion in t-SNE framework.

\goodbreak
\section*{Appendix B. Data Sources and Transcriptomic Data Preprocessing}

%\paragraph{Transcriptomic data sets}

The brain organoid data sets \citep{kanton2019organoid} were downloaded from \url{ https://www.ebi.ac.uk/arrayexpress/experiments/E-MTAB-7552/} in form of UMI  counts and metadata tables. The metadata table for the chimpanzee data set was taken from the supplementary materials of the original publication. We used gene counts mapped to the consensus genome, and selected all cells that passed quality control by the original authors (\texttt{in\_FullLineage=TRUE} in  metadata tables). For human organoid data, we only used cells from the \texttt{409b2} cell line, to simplify the analysis (the original publication combined cells from two cell lines and needed to perform batch correction).

The hydra data set (Figure~\ref{fig:hydra}; \citealp{siebert19hydra}) was downloaded in form of UMI counts from \url{https://www.ncbi.nlm.nih.gov/geo/query/acc.cgi?acc=GSE121617}.

\begin{sloppypar}
The zebrafish data set (Figure~\ref{fig:zfish}; \citealp{wagner2018zfish}) was downloaded in form of UMI counts from \url{https://kleintools.hms.harvard.edu/paper_websites/wagner_zebrafish_timecourse2018/WagnerScience2018.h5ad}.

The adult mouse cortex data set (Figure~\ref{fig:tasic}; \citealp{tasic2018shared}) was downloaded in form of read counts from \url{http://celltypes.brain-map.org/api/v2/well_known_file_download/694413985} and \url{http://celltypes.brain-map.org/api/v2/well_known_file_download/694413179} for the VISp and ALM cortical areas, respectively. Only exon counts were used here. The cluster labels and cluster colors were retrieved from \url{http://celltypes.brain-map.org/rnaseq/mouse/v1-alm}.
\end{sloppypar}

To preprocess each data set, we selected the 1000 most variable genes using the procedure from \citet{kobak2018art} with default parameters (for the mouse cortex data set we used 3000 genes and \texttt{threshold=32}; \citealp{kobak2018art}) and followed the preprocessing pipeline from the same paper: normalized all counts by cell sequencing depth (sum of gene counts in each cell), multiplied by the median cell depth (or 1 million in case of mouse cortex data), applied $\log_2(x+1)$ transformation, did PCA, and retained 50 leading PCs.

%\paragraph{MNIST-like data sets}
%\bigskip\noindent
The data sets shown in Figures~\ref{fig:faminst},~\ref{fig:kannada},~and~\ref{fig:kuzmnist} have been published explicitly to function as drop-in replacements for the hand-written MNIST data set.  The data set variants that we used here all consist of a total $n=70\,000$ images of $28\times 28$ pixels, in 10 balanced classes.  The input was preprocessed like the original MNIST data set by reducing the dimensions to~50 via PCA.  Fashion and Kuzushiji MNIST were downloaded via OpenML with the keys \texttt{Fashion-MNIST} (\url{https://www.openml.org/d/40996}) and \texttt{Kuzushiji-MNIST} (\url{https://www.openml.org/d/41982}), respectively.  Kannada MNIST was downloaded from \url{https://github.com/vinayprabhu/Kannada_MNIST}.

%\section*{Supplementary Tables}
%\renewcommand{\thetable}{S\arabic{table}}

\counterwithin{figure}{section}
\renewcommand{\thefigure}{A\arabic{figure}}
\setcounter{figure}{0}  
\section*{Appendix C. Supporting Experiments}

\vfill
\begin{figure}[h]
  \centering
  \includegraphics[width=\linewidth]{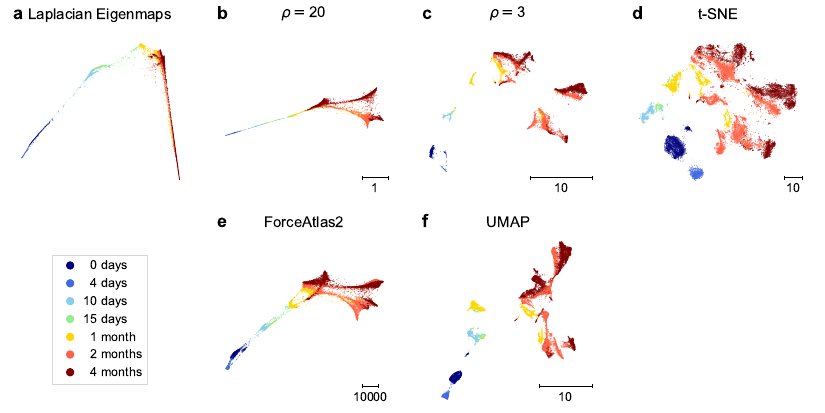}
  \caption{\textbf{Neighbor embeddings of the single-cell RNA-seq developmental data (human, high $k$).} 
  The same as Figure~\ref{fig:treutleinHuman}, but LE, FA2, and UMAP used $k=150$ (instead of our default $k=15$), while t-SNE used perplexity 300 (instead of our default 30).}
  \label{fig:treutleinHumanHighK}
\end{figure}
\vfill

\begin{figure}
  \centering
  \includegraphics[width=\linewidth]{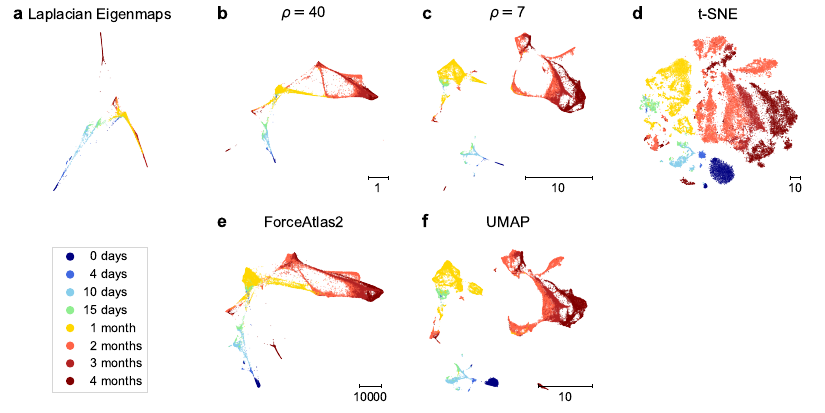}
  \caption{\textbf{Neighbor embeddings of the single-cell RNA-seq developmental data (chimpanzee).} Cells were sampled from chimpanzee brain organoids at eight time points between 0 days and 4 months into the development \citep{kanton2019organoid}. Sample size $n=36\,884$. Data were reduced with PCA to 50 dimensions. See Appendix B for transcriptomic data preprocessing steps.}
  \label{fig:treutleinChimp}
\end{figure}

\begin{figure}
  \centering
  \includegraphics[width=\linewidth]{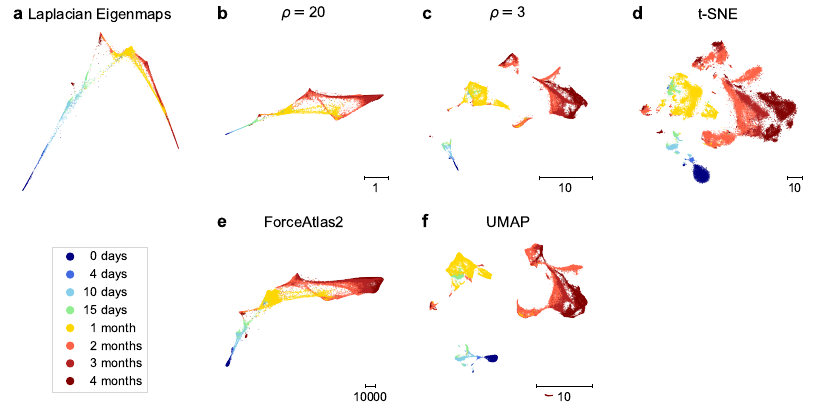}
  \caption{\textbf{Neighbor embeddings of the single-cell RNA-seq developmental data (chimpanzee, high $k$).} The same as Figure~\ref{fig:treutleinChimp}, but LE, FA2, and UMAP used $k=150$ (instead of our default $k=15$), while t-SNE used perplexity 300 (instead of our default 30).}
  \label{fig:treutleinChimpHighK}
\end{figure}

\begin{figure}
  \centering
  \includegraphics[width=\linewidth]{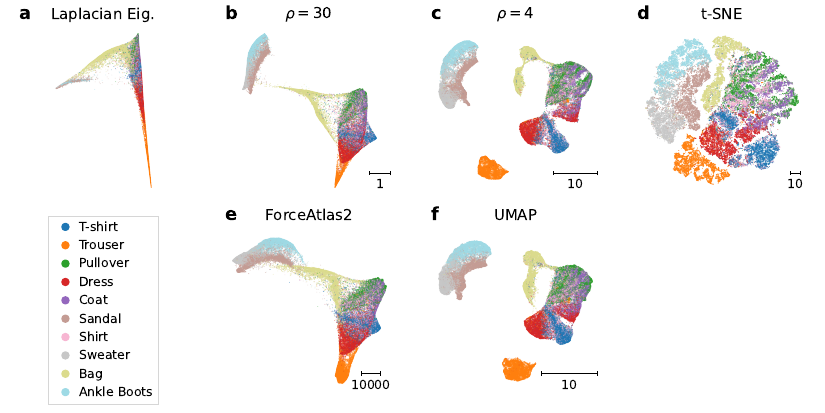}
  \caption{\textbf{Fashion MNIST data set \citep{xiao2017fashion}}. Sample size $n=70\,000$. Dimensionality was reduced to 50 with PCA. Colors correspond to 10 classes, see legend.}
  \label{fig:faminst}
\end{figure}

\begin{figure}
  \centering
  \includegraphics[width=\linewidth]{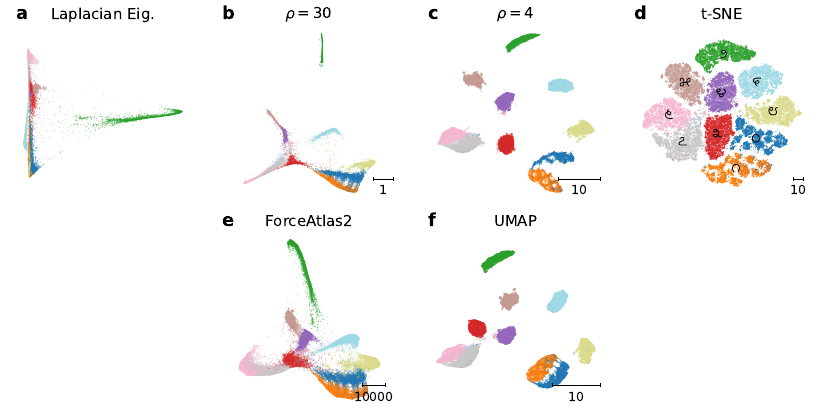}
  \caption{\textbf{Kannada MNIST data set \citep{prabhu2019kannada}.} Sample size $n=70\,000$. Dimensionality was reduced to 50 with PCA. Colors correspond to 10 Kannada digits shown in panel~(d).}
  \label{fig:kannada}
\end{figure}

\begin{figure}
  \centering
  \includegraphics[width=\linewidth]{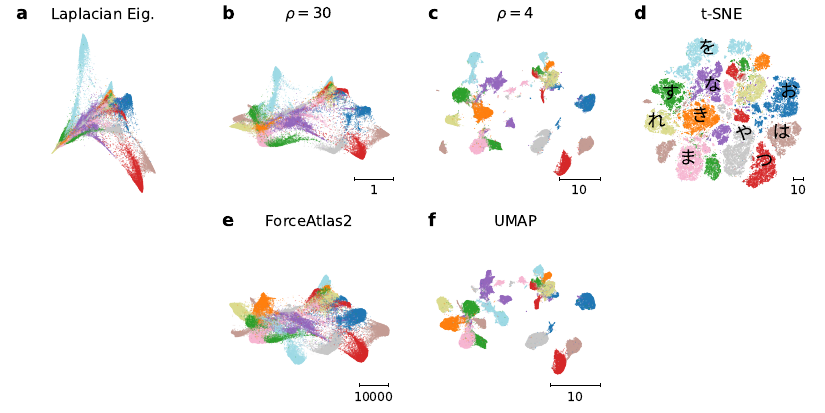}
  \caption{\textbf{Kuzushiji MNIST data set \citep{clanuwat2018kuzmnist}.} Sample size $n=70\,000$. Dimensionality was reduced to 50 with PCA. Colors correspond to 10 Kanji characters shown in panel (d).}
  \label{fig:kuzmnist}
\end{figure}

\begin{figure}
  \centering
  \includegraphics[width=\linewidth]{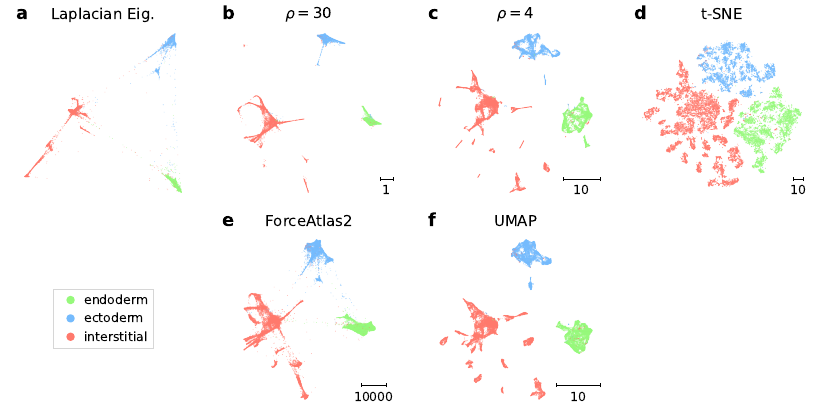}
  \caption{\textbf{Single-cell RNA-seq data of a hydra \citep{siebert19hydra}.} Sample size $n=24\,985$. Dimensionality was reduced to 50 with PCA. See Appendix B for transcriptomic data preprocessing steps. Color corresponds to cell classes.}
  \label{fig:hydra}
\end{figure}

\begin{figure}
  \centering
  \includegraphics[width=\linewidth]{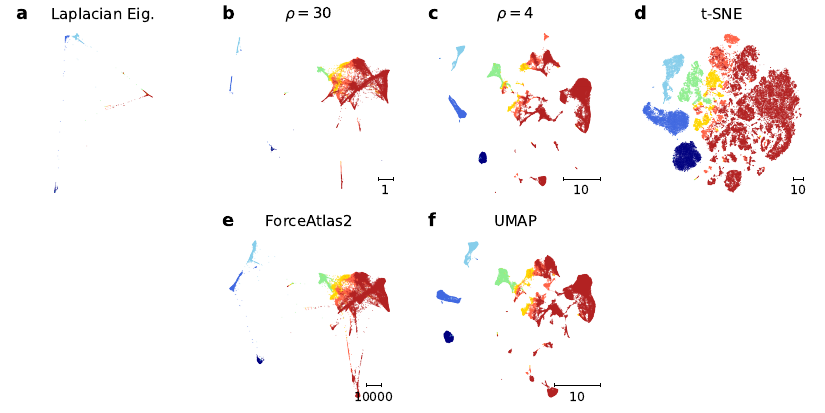}
  \caption{\textbf{Single-cell RNA-seq data of a zebrafish embryo \citep{wagner2018zfish}.} Sample size $n=63\,530$. Dimensionality was reduced to 50 with PCA. See Appendix B for transcriptomic data preprocessing steps. Color corresponds to the developmental stage, indicating the hours post fertilization (hpf).}
  \label{fig:zfish}
\end{figure}

\begin{figure}
  \centering
  \includegraphics[width=\linewidth]{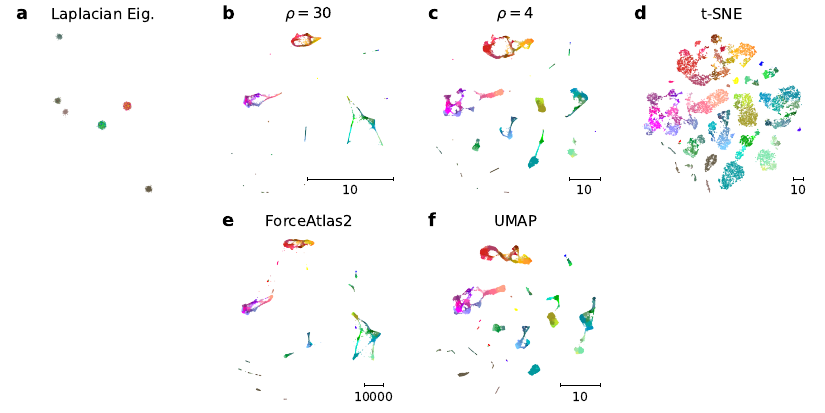}
  \caption{\textbf{Single-cell RNA-seq data of adult mouse cortex \citep{tasic2018shared}.} Sample size $n=23\,822$. Dimensionality was reduced to 50 with PCA. See Appendix B for transcriptomic data preprocessing steps. Colors are taken from the original publication (warm colors: inhibitory neurons; cold colors: excitatory neurons; grey/brown: non-neural cells).  We added Gaussian noise to the LE embedding in panel (a) to make the clusters more visible. In this data set, the $k$NN graph is disconnected and has 6 components, resulting in 6 distinct points in the LE embedding.}
  \label{fig:tasic}
\end{figure}

\begin{figure}[t]
  \centering
  \floatbox[{\capbeside\thisfloatsetup{capbesideposition={right,center},capbesidewidth=0.4\textwidth}}]{figure}[\FBwidth]
  {\includegraphics[width=\linewidth]{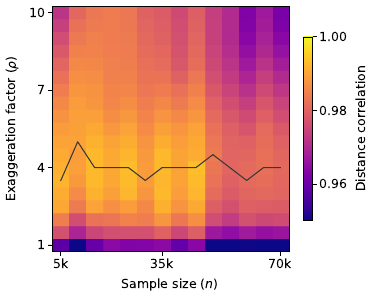}}
  {\caption{Distance correlations between t-SNE with $\rho\in[1,10]$ and UMAP depending on the sample size, for MNIST subsets of size $n\in [5\,000, 70\,000]$. Black line indicates best matching $\rho$ values.}
  \label{fig:layoutCorrelationsEtc}}
\end{figure}

\begin{figure}
  \centering
  \includegraphics[width=\linewidth]{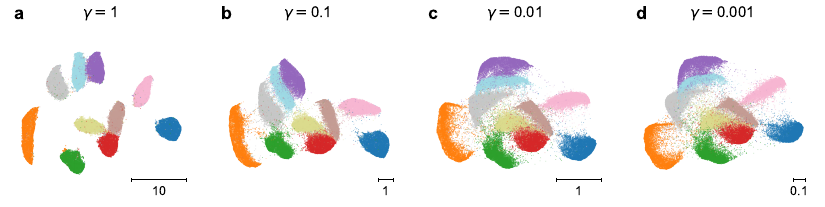}
  \caption{\textbf{Decreasing the repulsion in UMAP}. \textbf{(a)} UMAP embedding of MNIST with $\gamma=1$ (default). \textbf{(b--d)} Decreasing $\gamma$ produces the same effect as increasing the exaggeration $\rho$ in t-SNE. Values $\gamma>1$ are not shown because we could not achieve a well-converged embedding for $\gamma\gg 1$.}
  \label{fig:umapAttraction}
\end{figure}

\begin{figure}[h]
  \centering
  \includegraphics[width=\linewidth]{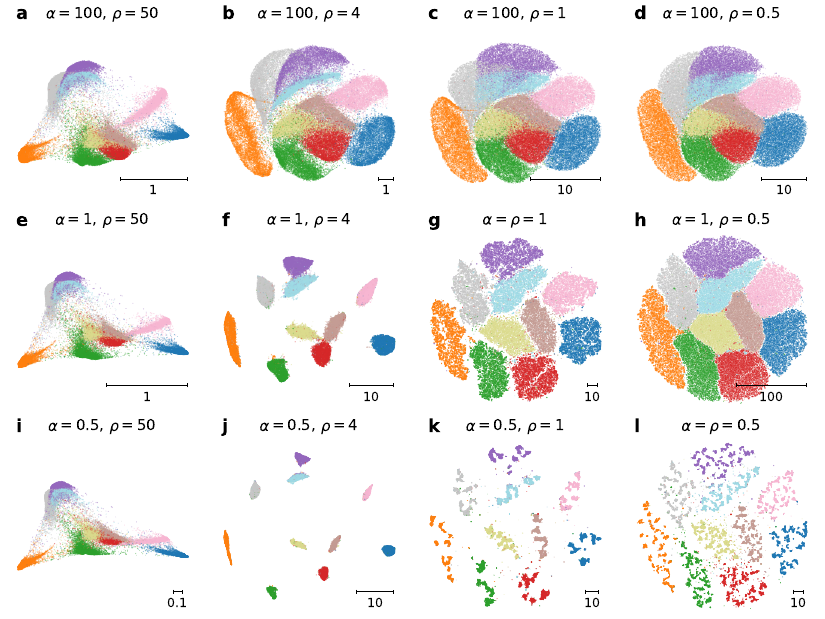}
  \caption{\textbf{Varying the tail-heaviness and exaggeration}.  Changes in the layout for t-SNE of the MNIST data set when varying the tail-heaviness \citep{kobak2019heavy,yang2009heavy} $\alpha\in\{100, 1, 0.5\}$ and the exaggeration factor~$\rho\in\{50, 4, 1, 0.5\}$.}
  \label{fig:tsne-tail-heavy}
\end{figure}

\begin{figure}
  \centering
  \includegraphics[width=\linewidth]{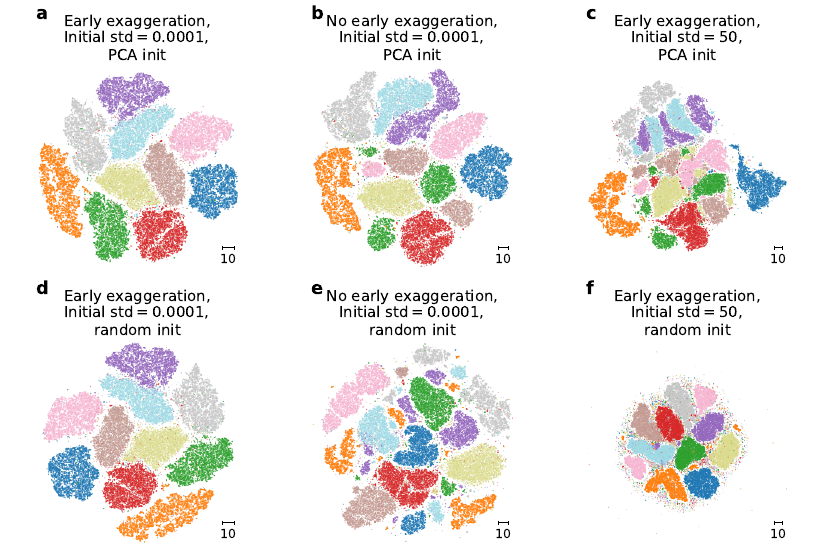}
  \caption{\textbf{The effect of early exaggeration on t-SNE}. \textbf{(a)} Default t-SNE embedding of MNIST. This uses early exaggeration and sets the standard deviation of PCA initialization to 0.0001. \textbf{(b)} T-SNE embedding without early exaggeration. This embedding is stuck in a suboptimal local minimum with some clusters split into multiple parts. \textbf{(c)} T-SNE embedding with early exaggeration, but with initial standard deviation set to 50. The attractive forces are too weak to pull the clusters together during the early exaggeration phase. \textbf{(d)} Default t-SNE with random initialization.  The cluster structure is recovered, but the placement of the clusters is different from (a). \textbf{(e)} Same experiment as in (b), but with random initialization.  The clusters are more fragmented due to less structure in the initialization and the lack of early exaggeration.  \textbf{(f)} Same experiment as (c), but with random initialization. Here again, the attractive forces are too weak to pull the clusters together, and in addition there are points on the periphery that got stuck there due to the large initial distances.}
  \label{fig:tsneParams}
\end{figure}

\clearpage
\bibliography{main}

\end{document}